\documentclass[9pt,twocolumn]{article}%[12pt, letterpaper]{article}
\usepackage[utf8]{inputenc}
\usepackage[T1]{fontenc}
\usepackage{comment}
\usepackage[english]{babel}
\usepackage{hyperref}
\usepackage{amssymb}
\usepackage{graphicx}
\usepackage{amsmath}
\usepackage{xcolor} % for inkscape exported figures
\usepackage{pgfbaseimage}
\usepackage{hhline} % for table
\usepackage{geometry}
\usepackage{enumitem}

\newgeometry{top=3.6cm,bottom=3.7cm}

\DeclareMathOperator*{\convone}{conv1D}

\title{
  \vspace{-10ex}
  Recurrent autoencoder with sequence-aware encoding
}

\author{Robert Susik \\
\href{mailto:rsusik@kis.p.lodz.pl}{rsusik@kis.p.lodz.pl} \\
\multicolumn{1}{p{.7\textwidth}}{\centering\emph{Institute of Applied Computer Science, \\Łódź University of Technology, Poland}}}
\date{2020}

\begin{document}

\twocolumn[
  \begin{@twocolumnfalse}
    \maketitle
    \begin{abstract}
Recurrent Neural Networks (RNN)
received a vast amount of attention
last decade. 
Recently, the architectures of Recurrent AutoEncoders (RAE)
found many applications in practice.
RAE can extract the semantically valuable information,
called context that represents a latent space useful for further processing.
Nevertheless, recurrent autoencoders are 
hard to train, and the training process takes much time.
In this paper, we propose an autoencoder architecture
with sequence-aware encoding,
which employs 1D convolutional layer to 
improve its performance in terms of model training time.
We prove that the recurrent 
autoencoder with sequence-aware encoding 
outperforms a standard RAE in terms of training speed
in most cases.
The preliminary results show that 
the proposed solution dominates over the standard RAE, 
and the training process is order of magnitude faster.
       \vspace{6ex}
    \end{abstract}
  \end{@twocolumnfalse}
]

\section{Introduction} \label{sec:intro}
Recurrent Neural Networks (RNN)
~\cite{RGW1986, W1990}
received a vast amount of attention
last decade and found a
wide range of applications such as 
language modelling~\cite{mikolov2011extensions, shi2019knowledge}, 
signal processing~\cite{ding2019wifi, shahtalebi2019training},
anomaly detection~\cite{nanduri2016anomaly, su2019robust}.

The RNN is (in short) a feedforward neural network adapted
to sequences of data that have the ability to 
map sequences to sequences
achieving excellent performance on time series.
Multiple layers of RNN can be stacked 
to process efficiently
long input sequences~\cite{graves2013generating, a81cdc63b81a42f5af92c81179c94532}.
The training process of deep recurrent neural network (DRNN)
is difficult because 
the gradients (in backpropagation through time~\cite{W1990}) 
either vanish or explode
~\cite{bengio1994learning, doya1993bifurcations}.
It means that despite the RNN can learn long
dependencies the training process may take
a much time or even fail.
The problem was resolved by
application of 
Long Short-Term Memory (LSTM)~\cite{hochreiter1997long}
or much newer and simpler
Gated Recurrent Units (GRU)~\cite{Cho2014b}.
Nevertheless, 
it is not easy to parallelise calculations 
in recurrent neural networks what impact the training time.

A different and efficient approach was proposed by
Aäron et al.~\cite{WaveNet2016} 
who
proved that stacked 1D convolutional 
layers can process efficiently long sequences
handling tens of thousands of times steps.
The CNNs have also been widely applied to autoencoder
architecture as a solution
for problems such as
outlier and anomaly detection~\cite{liao2018unified, kieu2019outlier, an2015variational}, 
noise reduction~\cite{chiang2019noise},  
and more.

Autoencoders~\cite{BK1988}
are unsupervised algorithms
trained to attempt to copy its input to its output.
The desirable side effect of this approach is
a latent representation (called \textit{context} or \textit{code}) 
of the input data.
The context is usually smaller than input data 
to extract only the semantically valuable information.
Encoder-Decoder (Sequence-to-Sequence)~\cite{cho2014learning, sutskever2014sequence}
architecture
looks very much like autoencoder and
consists of two blocks: encoder, and decoder, both 
containing a couple of RNN layers.
The encoder takes the input data
and generates the code (a semantic summary)
used to represent the input.
Later, the decoder processes the code
and generates the final output.
The encoder-decoder approach allows having variable-length
input and output sequences in contrast to
classic RNN solutions. 
There are several related attempts,
including an interesting approach introduced by Graves~\cite{graves2013generating}
that have been later successfully applied in practice in
~\cite{bahdanau2014neural, luong2015effective}.
The authors proposed a novel differentiable
attention mechanism that allows the decoder to focus on 
appropriate words at each time step.
This technique improved state of the art 
in neural machine translation (NMT) and was later 
applied even without any recurrent or convolutional 
layers~\cite{vaswani2017attention}.
Besides the machine translation, 
there are multiple variants and applications of 
the Recurrent AutoEncoders (RAE).
In~\cite{FA2014},
the proposed generative model 
of variational recurrent autoencoder (VRAE)
learns the latent vector representation of data 
and use it
to generate samples.
Another variational autoencoder was introduced
in~\cite{van2017neural, garbacea2019low}
where authors apply convolutional layers and WaveNet
for audio sequence.
Interesting approach,
the Feedback Recurrent AutoEncoder (FRAE)
was presented in~\cite{yang2020feedback}.
In short, the idea is to add
a connection that provides feedback from
decoder to encoder. 
This design allows efficiently compressing the sequences of speech spectrograms.

In this paper, we present an autoencoder architecture, 
which employs 1D convolutional layer in order to 
improve its performance in terms of training 
time and model accuracy.
We also propose a different interpretation of
the context (the final hidden state of the encoder). 
We transform the context into 
the sequence that is passed 
to the decoder.
This technical trick,
even without changing other elements of architecture,
improves the performance 
of recurrent autoencoder.

We demonstrate the power of the proposed architecture for 
time series reconstruction.
We perform a wide range of experiments on a dataset of generated signals, 
and the preliminary results are promising.

Following contributions of this work can be enumerated:
(i) We propose a recurrent autoencoder with sequence-aware encoding
that trains much faster than standard RAE.
(ii) We suggest an extension to proposed solution 
which employs the 1D convolutional layer
to make the solution more flexible.
(iii) We show that this architecture
performs very well on univariate and multivariate 
time series reconstruction.

\section{The model}
\label{sec:model}

In this section, we describe our approach
and its variants. 
We also discuss the advantages and disadvantages 
of the proposed architecture and
suggest possible solutions to its limitation.

\begin{figure}[h]
  \centerline{
    \def\svgwidth{0.8\linewidth}
    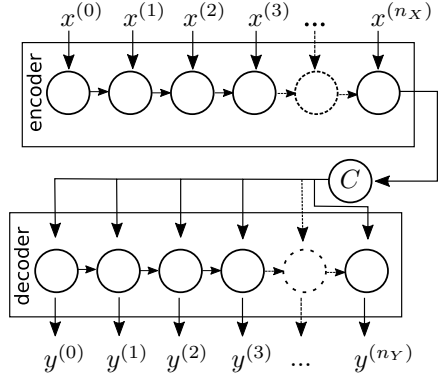
  }
  \caption{Recurrent AutoEncoder (RAE)~\cite{cho2014learning, sutskever2014sequence},
  called also Encoder-Decoder or Sequence to Sequence.}
  \label{fig:classic_arch}
\end{figure}

The recurrent autoencoder
generates an output sequence
$Y = (y^{(0)}, y^{(1)}, \ldots, y^{(n_Y-1)})$
for given an input sequence
$X = (x^{(0)}, x^{(1)}, \ldots, x^{(n_X-1)})$,
where $n_Y$ and $n_X$ are the sizes 
of output and input sequences respectively
(both can be of the same or different size).
Usually, $X = Y$ to force autoencoder learning
the semantic meaning of data.
First, the input sequence is encoded by
the RNN encoder, and then the given 
fixed-size context variable 
$C$ 
is decoded by the decoder (usually also RNN),
see Figure~\ref{fig:classic_arch}.

\subsection{Recurrent AutoEncoder with Sequential context (RAES)} 
\label{ssec:model_basic}

\begin{figure}[h]
  \centerline{
    \def\svgwidth{0.75\linewidth}
    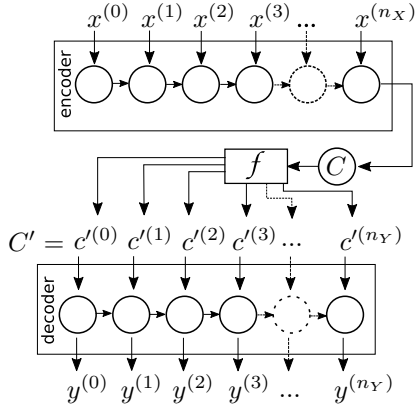
  }
  \caption{Recurrent AutoEncoder with Sequence-aware encoding (RAES).}
  \label{fig:basic_arch}
\end{figure}
We propose a recurrent autoencoder 
architecture (Figure~\ref{fig:basic_arch})
where the context $C$
(the output of the final hidden state of the encoder)
is interpreted as the sequence $C'$,
of $m_{C'} = \lambda$ features
producing the output sequence $Y$.
The 
$C = (c_i)^{n_C-1}_{i=0}$
is transformed to
\begin{equation}
  C' = ((c_{i\lambda+j})^{\lambda - 1}_{j=0})^{n_C/\lambda - 1}_{i=0}
\end{equation}
where $\lambda = n_C / n_X$ ($\lambda \in \mathbb{N}$).

Once the context is transformed 
($C' = (c'^{(0)}, c'^{(1)}, \ldots, c'^{(n_X-1)})$),
the decoder starts to decode the sequence $C'$
of $m_{C'} = \lambda$ features.
This technical trick
in the data structure
speeds up the training process (Section~\ref{sec:exp}).
Additionally, 
this way, we put some sequential meaning to the context.
The one easily solvable disadvantage of this 
solution is the fact that the size of 
context must be multiple of input
sequence length $n_C = \lambda n_X$,
where $n_C$ is the size of context $C$.

\subsection{RAES with 1D Convolutional layer (RAESC)}
\label{ssec:model_conv1D}

\begin{figure}[h]
  \centerline{
    \def\svgwidth{0.8\linewidth}
    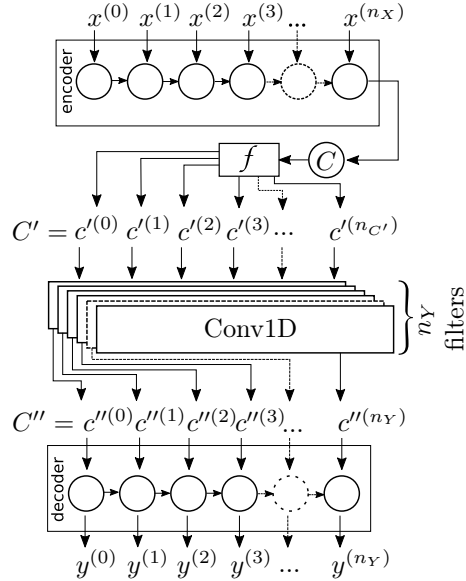
  }
  \caption{Recurrent AutoEncoder with Sequence-aware encoding and 1D Convolutional layer (RAESC).}
  \label{fig:conv_arch}
\end{figure}
In order to solve the limitation mentioned
in the previous section (Section~\ref{ssec:model_basic}), 
we propose to add 1D convolutional layer (and max-pooling layer)
to the architecture 
right before the decoder (Figure~\ref{fig:conv_arch}).
This approach gives the ability to control
the number of output channels 
(also denoted as \textit{feature detectors} or \textit{filters}),
defined as follows:
\begin{equation}
  C''(i) = \sum_{k}\sum_{l}C'(i + k, l)w(k, l)
\end{equation}
In this case,
the $n_C$ does not have to be multiple of $n_X$, 
thus to have the desired output sequence of $n_Y$ length,
the number of filters should be equal to $n_Y$. 
Moreover, the output of the 1D convolution layer 
$C'' = \convone(C')$
should be transposed, 
hence each channel becomes an element of
the sequence as shown in Figure~\ref{fig:conv_arch}.
Finally, the desired number of features on output $Y$ 
can be configured with hidden state size of the decoder.

A different and simpler approach to 
solve the mentioned limitation is
stretching the $C$ to the size of decoder input
and filling in the gaps with averages.

The described variant is very simplified and 
is only an outline of proposed
recurrent autoencoder architecture 
(the middle part of it, to be more precise)
which 
can be extended by adding pooling
and recurrent layers or
using different convolution parameters
(such as stride, or dilation values).
Furthermore, in our view, this approach
could be easily applied to other 
RAE architectures
(such as~\cite{yang2020feedback, garbacea2019low}).

\section{Experiments} \label{sec:exp}
In order to evaluate the proposed approach, 
we run a few experiments,
using a generated dataset of signals.
We tested the following algorithms:
\begin{itemize}[leftmargin=*,align=left]
  \setlength\itemsep{0em}
  \item Standard Recurrent AutoEncoder (RAE)~\cite{cho2014learning, sutskever2014sequence}.
  \item RAE with Sequence-aware encoding (RAES).
  \item RAES with Convolutional and max-pooling layer (RAESC).
\end{itemize}

The structure of decoder and encoder is the same
in all algorithms.
Both, decoder and encoder are single GRU~\cite{Cho2014b} layer,
with additional time distributed fully connected layer 
in the output of decoder.
The algorithms were implemented in Python 3.7.4
with TensorFlow 2.3.0.
The experiments
were run on a desktop PC 
with an Intel i5-3570 CPU clocked at 3.4 GHz with 256 KB
L1, 1 MB L2 and 6 MB L3 cache and GTX 1070 graphic card. 
The test machine was equipped with 16 GB of
1333 MHz DDR3 RAM and running Fedora 28 64-bit OS.
The dataset contains 5000 sequences of size 200
with \{1, 2, 4, 8\} features.
The dataset was shuffled and split to training 
and validation sets in proportions of 80:20,
respectively.
We trained the models with Adam optimizer~\cite{kingma2014adam}
in batches of size 100 and 
Mean Squared Error (MSE) loss function.

In the first set of analyses we investigated
the impact of context size and
the number of features on performance.
We noticed that there is a considerable difference
in training speed (number of epochs needed to 
achieve plateau) between the classic approach 
and ours.
To prove whether our approach has an advantage over the standard RAE, 
we performed tests with different size of the context $n_C$ 
and a different number of input features $m_X$. 
We set the $n_C$ size proportionally 
to the size of the input and we denote it as:
\begin{equation}
  \sigma = \frac{n_C}{m_Xn_X}  
\end{equation}

\begin{figure}[h!]
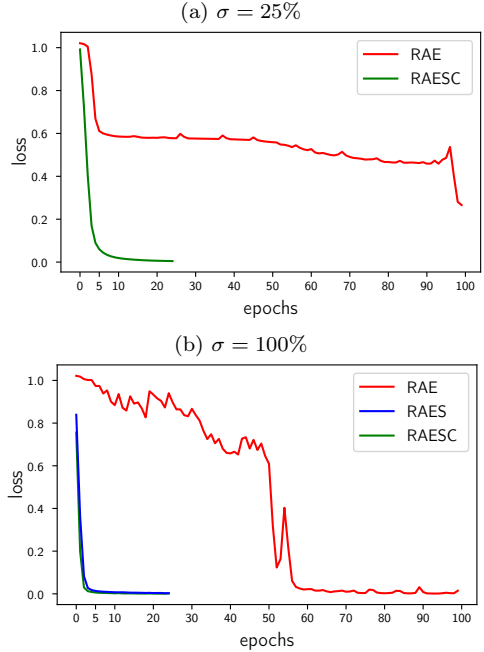

  \centering{
      \footnotesize
      (a) $\sigma = 25\%$\\
      \resizebox{0.9\columnwidth}{!}{\input{fig3a.pgf}} \\
      (b) $\sigma = 100\%$\\
      \resizebox{0.9\columnwidth}{!}{\input{fig3b.pgf}}
  }
  \caption{Loss as function of epoch number for univariate data and $\sigma = \{25\%, 100\%\}$.}
  \label{fig:result_f1_h}
\end{figure}

Figure~\ref{fig:result_f1_h} proves 
that the training of standard RAE 
takes much more time (epochs) than RAESC.
In chart a) 
the size of context is set to $\sigma = 25\%$ and 
in b) it is set to $\sigma = 100\%$
of the input size.
For $\sigma = 25\%$ the RASEC achieves plateau after 
20 epochs while standard RAE does not at all
(it starts decreasing after nearly 100 epochs). % TODO: starts decresing?
There is no RAES result presented in this plot
because of the limitation mentioned in Section~\ref{ssec:model_basic} 
(size of code was too small to fit 
the output sequence length).
For the $\sigma = 100\%$ both RASEC and RAES achieve the plateau 
in less than five epochs while the standard RAE 
after 50 epochs (order of magnitude faster).

\begin{figure}[h!]
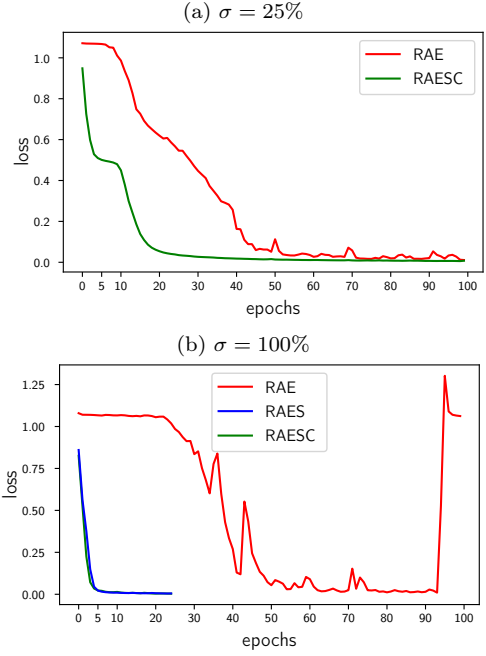

  \centering{
      \footnotesize
      (a) $\sigma = 25\%$\\
      \resizebox{0.9\columnwidth}{!}{\input{fig4a.pgf}} \\
      (b) $\sigma = 100\%$\\
      \resizebox{0.9\columnwidth}{!}{\input{fig4b.pgf}}
  }
  \caption{
    Loss as function of epoch number for two features ($m_X = 2$) and $\sigma = \{25\%, 100\%\}$.}
  \label{fig:result_f2_h}
\end{figure}

Figure~\ref{fig:result_f2_h} shows the loss in function of 
the number of epochs for two features in input data.
This experiment confirms that both RAES and RAESC dominates
in terms of training speed, but a slight difference can be noticed
in comparison to univariate data (Figure~\ref{fig:basic_arch}).
It shows that
the RAE achieves plateau in \~50 epochs
for both cases 
while RAES and RAESC after 20 epochs for $\sigma = 25\%$ and 
in about five epochs for $\sigma = 100\%$.

\begin{figure}[h!]
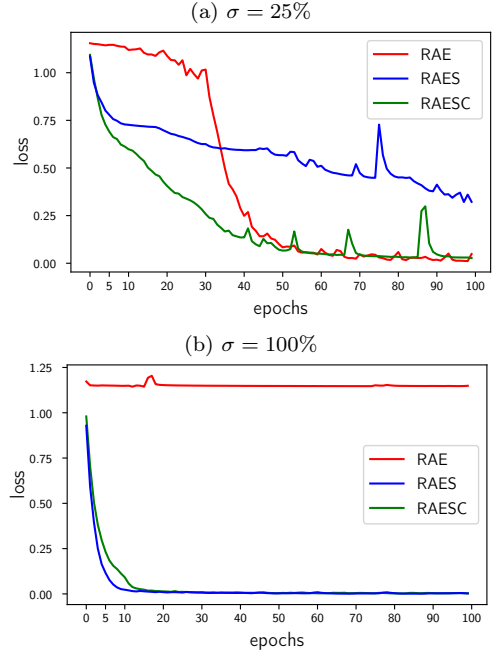

  \centering{
      \footnotesize
      (a) $\sigma = 25\%$\\
      \resizebox{0.9\columnwidth}{!}{\input{fig5a.pgf}} \\
      (b) $\sigma = 100\%$\\
      \resizebox{0.9\columnwidth}{!}{\input{fig5b.pgf}}
  }
  \caption{Loss as function of epoch number for $m_X = 8$ and $\sigma = \{25\%, 100\%\}$.}
  \label{fig:result_f8_h}
\end{figure}

In Figure~\ref{fig:result_f8_h}
is presented loss in function of the number of epochs 
for 8 features.
This figure is interesting in several ways
if compared to the previous ones (Figures~\ref{fig:result_f1_h}, \ref{fig:result_f2_h}).
The chart a) shows that,
for much larger number of features and relatively small size of the context, 
the training time of RAES variant 
is much longer.
The similar observation may be noticed for RAESC, 
where the loss drops much faster 
than the standard RAE at the begining of the training,
but achieves the plateau at almost the same step.
On the other hand, chart b) shows that for larger 
size of context, the proposed solution dominates.
The most striking fact to emerge from these results is that 
the RAE does not drop in the whole period.

\begin{figure}[h!]
  \centering{
      \footnotesize
      \resizebox{0.9\columnwidth}{!}{\input{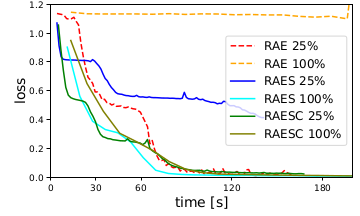}} \\
  }
  \caption{Loss as function of time [s] for $m_X = 4$ and $\sigma = \{25\%, 100\%\}$.}
  \label{fig:result_f8_h_tuneing}
\end{figure}

We compared also the algorithms' performance for different context size
($\sigma$).
Figure~\ref{fig:result_f8_h_tuneing} 
presents loss in a function of time (in seconds) for 4 features.
Each variant was tested for 100 epochs 
with limited time to 200 seconds.
As expected, 
we can clearly see that generally the proposed solution
converges faster than the RAE.
Nevertheless,
the RAEs loss function for $\sigma = 25\%$ 
is very similar to RAES after about one minute.
For $\sigma = 100\%$ 
the loss of RAE does not drop for the whole period
and even raise near 200 seconds.
The RAEs' result could be much worse for univariate data
or a smaller number of features.

\begin{table}[!htb]
\begin{center}
\resizebox{\linewidth}{!}{%
\begin{tabular}{|c|l||r|r|r|}
\hline
features    &              & \multicolumn{3}{c|}{$\sigma$} \\\cline{3-5}
($m_X$)     & algorithm    & 25\%    & 50\%     & 100\% \\
\hhline{|=|=#=|=|=|}
~           & RAE        & 1.05    & 1.00     & 1.41  \\
1           & RAES         &  -~~    &  -~~     & 1.23  \\
~           & RAESC        & 0.97    & 0.98     & 1.64  \\
\hline
~           & RAE        & 1.00    & 1.47     & 3.69  \\
2           & RAES         &  -~~    & 1.29     & 3.10  \\
~           & RAESC        & 0.97    & 1.63     & 3.99  \\
\hline
~           & RAE        & 1.47    & 3.66     & 10.60 \\
4           & RAES         & 1.31    & 3.08     & 8.24  \\
~           & RAESC        & 1.60    & 3.97     & 10.95 \\
\hline
~           & RAE        & 3.65    & 10.51    & 35.01 \\
8           & RAES         & 3.14    & 8.28     & 26.28 \\
~           & RAESC        & 3.89    & 10.80    & 35.68 \\
\hline
\end{tabular}
}
\end{center}
\caption{Epoch time [s] (median) for different number of features ($m_X$) and context size ($\sigma$).}
\label{tab:time}
\end{table}

Finally, we measured the training time of each algorithm
to confirm that proposed solution converges faster than standard RAE
for the same size of context.
The Table~\ref{tab:time} shows the median of epoch time 
for different number of features and context size. 
The table proves that the RAES is 
14\% faster than RAE for univariate data and 
about 33\% faster than RAE for $m_X = 8$.
The training of RAESC algorithm
takes a slightly more time than RAE, 
which is marginal (less than 2\%) for $m_X = 8$.

\section{\hspace*{-3pt}Conclusions~and~future~work}
In this work, we proposed an
autoencoder with sequence-aware encoding.
We proved that this solution 
outperforms a standard RAE in terms of training speed
(for the same size of context)
in most cases.

The experiments confirmed
that the training of proposed architecture
is much faster than the standard RAE.
The context size and a number of features in the input sequence 
have a high impact on training performance.
Only for relatively large number of features and small size of the context
the proposed solution achieves comparable results to standard RAE.
In other cases our solution dominates and the training time
is order of magnitude shorter.

In our view these results constitute a good 
initial step toward further research.
The proposed architecture was much 
simplified and the use of different layers
or hyperparameter tunning seems to offer 
great opportunities.
We belive that the proposed solution has 
a wide range of practical applications
and it is worth confirming.

\bibliographystyle{abbrv}
\bibliography{main}

\begin{thebibliography}{10}

\bibitem{an2015variational}
J.~An and S.~Cho.
\newblock Variational autoencoder based anomaly detection using reconstruction
  probability.
\newblock {\em Special Lecture on IE}, 2(1):1--18, 2015.

\bibitem{bahdanau2014neural}
D.~Bahdanau, K.~Cho, and Y.~Bengio.
\newblock Neural machine translation by jointly learning to align and
  translate.
\newblock {\em arXiv preprint arXiv:1409.0473}, 2014.

\bibitem{bengio1994learning}
Y.~Bengio, P.~Simard, and P.~Frasconi.
\newblock Learning long-term dependencies with gradient descent is difficult.
\newblock {\em IEEE transactions on neural networks}, 5(2):157--166, 1994.

\bibitem{BK1988}
H.~Bourlard and Y.~Kamp.
\newblock Auto-association by multilayer perceptrons and singular value
  decomposition.
\newblock {\em Biological cybernetics}, 59(4-5):291--294, 1988.

\bibitem{chiang2019noise}
H.-T. Chiang, Y.-Y. Hsieh, S.-W. Fu, K.-H. Hung, Y.~Tsao, and S.-Y. Chien.
\newblock Noise reduction in ecg signals using fully convolutional denoising
  autoencoders.
\newblock {\em IEEE Access}, 7:60806--60813, 2019.

\bibitem{Cho2014b}
K.~Cho, B.~van Merrienboer, D.~Bahdanau, and Y.~Bengio.
\newblock On the properties of neural machine translation: Encoder-decoder
  approaches.
\newblock {\em CoRR}, abs/1409.1259, 2014.

\bibitem{cho2014learning}
K.~Cho, B.~Van~Merri{\"e}nboer, C.~Gulcehre, D.~Bahdanau, F.~Bougares,
  H.~Schwenk, and Y.~Bengio.
\newblock Learning phrase representations using rnn encoder-decoder for
  statistical machine translation.
\newblock {\em arXiv preprint arXiv:1406.1078}, 2014.

\bibitem{ding2019wifi}
J.~Ding and Y.~Wang.
\newblock Wifi csi-based human activity recognition using deep recurrent neural
  network.
\newblock {\em IEEE Access}, 7:174257--174269, 2019.

\bibitem{doya1993bifurcations}
K.~Doya.
\newblock Bifurcations of recurrent neural networks in gradient descent
  learning.
\newblock {\em IEEE Transactions on neural networks}, 1(75):218, 1993.

\bibitem{FA2014}
O.~Fabius and J.~R. van Amersfoort.
\newblock Variational recurrent auto-encoders.
\newblock {\em arXiv preprint arXiv:1412.6581}, 2014.

\bibitem{garbacea2019low}
C.~G{\^a}rbacea, A.~van~den Oord, Y.~Li, F.~S. Lim, A.~Luebs, O.~Vinyals, and
  T.~C. Walters.
\newblock Low bit-rate speech coding with vq-vae and a wavenet decoder.
\newblock In {\em ICASSP 2019-2019 IEEE International Conference on Acoustics,
  Speech and Signal Processing (ICASSP)}, pages 735--739. IEEE, 2019.

\bibitem{graves2013generating}
A.~Graves.
\newblock Generating sequences with recurrent neural networks.
\newblock {\em arXiv preprint arXiv:1308.0850}, 2013.

\bibitem{hochreiter1997long}
S.~Hochreiter and J.~Schmidhuber.
\newblock Long short-term memory.
\newblock {\em Neural computation}, 9(8):1735--1780, 1997.

\bibitem{kieu2019outlier}
T.~Kieu, B.~Yang, C.~Guo, and C.~S. Jensen.
\newblock Outlier detection for time series with recurrent autoencoder
  ensembles.
\newblock In {\em IJCAI}, pages 2725--2732, 2019.

\bibitem{kingma2014adam}
D.~P. Kingma and J.~Ba.
\newblock Adam: A method for stochastic optimization.
\newblock {\em arXiv preprint arXiv:1412.6980}, 2014.

\bibitem{liao2018unified}
W.~Liao, Y.~Guo, X.~Chen, and P.~Li.
\newblock A unified unsupervised gaussian mixture variational autoencoder for
  high dimensional outlier detection.
\newblock In {\em 2018 IEEE International Conference on Big Data (Big Data)},
  pages 1208--1217. IEEE, 2018.

\bibitem{luong2015effective}
M.-T. Luong, H.~Pham, and C.~D. Manning.
\newblock Effective approaches to attention-based neural machine translation.
\newblock {\em arXiv preprint arXiv:1508.04025}, 2015.

\bibitem{mikolov2011extensions}
T.~Mikolov, S.~Kombrink, L.~Burget, J.~{\v{C}}ernock{\`y}, and S.~Khudanpur.
\newblock Extensions of recurrent neural network language model.
\newblock In {\em 2011 IEEE international conference on acoustics, speech and
  signal processing (ICASSP)}, pages 5528--5531. IEEE, 2011.

\bibitem{nanduri2016anomaly}
A.~Nanduri and L.~Sherry.
\newblock Anomaly detection in aircraft data using recurrent neural networks
  (rnn).
\newblock In {\em 2016 Integrated Communications Navigation and Surveillance
  (ICNS)}, pages 5C2--1. Ieee, 2016.

\bibitem{a81cdc63b81a42f5af92c81179c94532}
R.~Pascanu, C.~Gulcehre, K.~Cho, and Y.~Bengio.
\newblock How to construct deep recurrent neural networks.
\newblock In {\em Proceedings of the Second International Conference on
  Learning Representations (ICLR 2014)}, 2014.

\bibitem{RGW1986}
D.~E. Rumelhart, G.~E. Hinton, and R.~J. Williams.
\newblock Learning representations by back-propagating errors.
\newblock {\em nature}, 323(6088):533--536, 1986.

\bibitem{shahtalebi2019training}
S.~Shahtalebi, S.~F. Atashzar, R.~V. Patel, and A.~Mohammadi.
\newblock Training of deep bidirectional rnns for hand motion filtering via
  multimodal data fusion.
\newblock In {\em GlobalSIP}, pages 1--5, 2019.

\bibitem{shi2019knowledge}
Y.~Shi, M.-Y. Hwang, X.~Lei, and H.~Sheng.
\newblock Knowledge distillation for recurrent neural network language modeling
  with trust regularization.
\newblock In {\em ICASSP 2019-2019 IEEE International Conference on Acoustics,
  Speech and Signal Processing (ICASSP)}, pages 7230--7234. IEEE, 2019.

\bibitem{su2019robust}
Y.~Su, Y.~Zhao, C.~Niu, R.~Liu, W.~Sun, and D.~Pei.
\newblock Robust anomaly detection for multivariate time series through
  stochastic recurrent neural network.
\newblock In {\em Proceedings of the 25th ACM SIGKDD International Conference
  on Knowledge Discovery \& Data Mining}, pages 2828--2837, 2019.

\bibitem{sutskever2014sequence}
I.~Sutskever, O.~Vinyals, and Q.~V. Le.
\newblock Sequence to sequence learning with neural networks.
\newblock In {\em Advances in neural information processing systems}, pages
  3104--3112, 2014.

\bibitem{WaveNet2016}
A.~van~den Oord, S.~Dieleman, H.~Zen, K.~Simonyan, O.~Vinyals, A.~Graves,
  N.~Kalchbrenner, A.~Senior, and K.~Kavukcuoglu.
\newblock Wavenet: A generative model for raw audio.
\newblock In {\em Arxiv}, 2016.

\bibitem{van2017neural}
A.~Van Den~Oord, O.~Vinyals, et~al.
\newblock Neural discrete representation learning.
\newblock In {\em Advances in Neural Information Processing Systems}, pages
  6306--6315, 2017.

\bibitem{vaswani2017attention}
A.~Vaswani, N.~Shazeer, N.~Parmar, J.~Uszkoreit, L.~Jones, A.~N. Gomez,
  {\L}.~Kaiser, and I.~Polosukhin.
\newblock Attention is all you need.
\newblock In {\em Advances in neural information processing systems}, pages
  5998--6008, 2017.

\bibitem{W1990}
P.~J. Werbos.
\newblock Backpropagation through time: what it does and how to do it.
\newblock {\em Proceedings of the IEEE}, 78(10):1550--1560, 1990.

\bibitem{yang2020feedback}
Y.~Yang, G.~Sauti{\`e}re, J.~J. Ryu, and T.~S. Cohen.
\newblock Feedback recurrent autoencoder.
\newblock In {\em ICASSP 2020-2020 IEEE International Conference on Acoustics,
  Speech and Signal Processing (ICASSP)}, pages 3347--3351. IEEE, 2020.

\end{thebibliography}

\end{document}